\documentclass[conference]{IEEEtran}
\IEEEoverridecommandlockouts

\usepackage{cite}
\usepackage{amsmath,amssymb,amsfonts}
\usepackage{float}
\usepackage{graphicx}
\usepackage{textcomp}
\usepackage[dvipsnames]{xcolor}
\def\BibTeX{{\rm B\kern-.05em{\sc i\kern-.025em b}\kern-.08em
    T\kern-.1667em\lower.7ex\hbox{E}\kern-.125emX}}

    \usepackage{graphicx}
\usepackage{epstopdf}
\usepackage{float}
\usepackage{bm,upgreek}
\usepackage{amsfonts}
\usepackage{enumitem}
\usepackage{mathtools}
\usepackage{multirow}
\usepackage{booktabs}
\usepackage[subnum]{cases}
\usepackage{setspace}
\usepackage{color}
\usepackage{blkarray, bigstrut} 
\usepackage[labelformat=simple]{caption}
\usepackage{subcaption}
\usepackage{xcolor}
\usepackage{pgfplots}
\pgfplotsset{compat=1.5,
  tick label style={font=\small},
  label style={font=\small}}
\usepackage{pgfplotstable}
\usepgfplotslibrary{statistics}
\usepackage{tikz}

\usepackage{pgfplots}
\usepackage{pgfplotstable}

\pgfplotsset{compat=1.7}
\usepgfplotslibrary{groupplots}

\usepackage[toc,acronym]{glossaries}
\usepackage{mathtools,nccmath}
\usepackage[ruled]{algorithm}
\usepackage{algpseudocode}

\DeclareMathOperator{\pred}{Pred}
\DeclareMathOperator{\layerv}{Layer^v}
\DeclareMathOperator{\layerf}{Layer^f}

\DeclareMathOperator{\updateFN}{Update}
\DeclareMathOperator{\aggregateFN}{Aggregate}

\algnewcommand{\LineComment}[1]{\State \(\#\) #1}

\pgfplotsset{
    grid style={
        dotted,
        gray
    },
}

\newenvironment{customlegend}[1][]{%
    \begingroup
    \csname pgfplots@init@cleared@structures\endcsname
    \pgfplotsset{#1}%
}{%
    \csname pgfplots@createlegend\endcsname
    \endgroup
}%

\def\addlegendimage{\csname pgfplots@addlegendimage\endcsname}

\makeatletter 
\newcommand{\linebreakand}{%
  \end{@IEEEauthorhalign}
  \hfill\mbox{}\par
  \mbox{}\hfill\begin{@IEEEauthorhalign}
}
\makeatother 

\begin{document}

\title{Scalability and Sample Efficiency Analysis of Graph Neural Networks for Power System State Estimation\\

\thanks{This paper has received funding from the European Union's Horizon 2020 research and innovation programme under Grant Agreement number 856967.}
}

\author{
\IEEEauthorblockN{Ognjen Kundacina}
\IEEEauthorblockA{\textit{The Institute for Artificial Intelligence } \\
\textit{Research and Development of Serbia }\\
Novi Sad, Serbia \\
ognjen.kundacina@ivi.ac.rs}
\and
\IEEEauthorblockN{Gorana Gojic}
\IEEEauthorblockA{\textit{The Institute for Artificial Intelligence } \\
\textit{Research and Development of Serbia }\\
Novi Sad, Serbia \\
gorana.gojic@ivi.ac.rs}
\and
\IEEEauthorblockN{Mirsad Cosovic}
\IEEEauthorblockA{\textit{Faculty of Electrical
Engineering} \\
\textit{University of Sarajevo}\\
Sarajevo, Bosnia and Herzegovina \\
mcosovic@etf.unsa.ba}
\and

\linebreakand

\IEEEauthorblockN{Dragisa Miskovic}
\IEEEauthorblockA{\textit{The Institute for Artificial Intelligence } \\
\textit{Research and Development of Serbia }\\
Novi Sad, Serbia \\
dragisa.miskovic@ivi.ac.rs}
\and
\IEEEauthorblockN{Dejan Vukobratovic}
\IEEEauthorblockA{\textit{Faculty of Technical Sciences} \\
\textit{University of Novi Sad}\\
Novi Sad, Serbia \\
dejanv@uns.ac.rs}
}

\maketitle

\begin{abstract}
Data-driven state estimation (SE) is becoming increasingly important in modern power systems, as it allows for more efficient analysis of system behaviour using real-time measurement data. This paper thoroughly evaluates a phasor measurement unit-only state estimator based on graph neural networks (GNNs) applied over factor graphs. To assess the sample efficiency of the GNN model, we perform multiple training experiments on various training set sizes. Additionally, to evaluate the scalability of the GNN model, we conduct experiments on power systems of various sizes. Our results show that the GNN-based state estimator exhibits high accuracy and efficient use of data. Additionally, it demonstrated scalability in terms of both memory usage and inference time, making it a promising solution for data-driven SE in modern power systems.
\end{abstract}


\section{Introduction}

\textbf{Motivation and literature review:} The state estimation (SE) algorithm is a key component of the energy management system that provides an accurate and up-to-date representation of the current state of the power system. Its purpose is to estimate complex bus voltages using available measurements, power system parameters, and topology information \cite{monticelli2000SE}. In this sense, the SE can be seen as a problem of solving large, noisy, sparse, and generally nonlinear systems of equations. The measurement data used by the SE algorithm usually come from two sources: the supervisory control and data acquisition (SCADA) system and the wide area monitoring system (WAMS) system. The SCADA system provides low-resolution measurements that cannot capture system dynamics in real-time, while the WAMS system provides high-resolution data from phasor measurement units (PMUs) that enable real-time monitoring of the system. The SE problem that considers measurement data from both WAMS and SCADA systems is formulated in a nonlinear way and solved in a centralized manner using the Gauss-Newton method \cite{monticelli2000SE}. On the other hand, the SE problem that considers only PMU data provided by WAMS has a linear formulation, enabling faster, non-iterative solutions.

In this work, we will focus on the SE considering only phasor measurements, described with a system of linear equations \cite{gol2014LinearSE}, which is becoming viable with the increasing deployment of PMUs. This formulation is usually solved using linear weighted least-squares (WLS), which involve matrix factorizations and can be numerically sensitive \cite{manousakis}. To address the numerical instability issues that often arise when using traditional SE solvers, researchers have turned to data-driven deep learning approaches \cite{zhang2019, zamzam2019}. These approaches, when trained on relevant datasets, are able to provide solutions even when traditional methods fail. For example, in \cite{zhang2019}, a combination of feed-forward and recurrent neural networks was used to predict network voltages using historical measurement data. In the nonlinear SE formulation, the study \cite{zamzam2019} demonstrates the use of deep neural networks as fast and quality initializers of the Gauss-Newton method.

Both linear WLS and common deep learning SE methods at its best approach quadratic computational complexity regarding the power system size. To fully utilize high sampling rates of PMUs, there is a motivation to develop SE algorithms with a linear computational complexity. One way of achieving this could be using increasingly popular graph neural networks (GNNs) \cite{GraphRepresentationLearningBook, velickovic2018graph}. GNNs have several advantages when used in power systems, such as permutation invariance, the ability to handle varying power system topologies, and requiring fewer trainable parameters and less storage space compared to conventional deep learning methods. One of the key benefits of GNNs is the ability to perform distributed inference using only local neighbourhood measurements, which can be efficiently implemented using the emerging 5G network communication infrastructure and edge computing \cite{kundacina5G2022}. This allows for real-time and low-latency decision-making even in large-scale networks, as the computations are performed at the edge of the network, closer to the data source, reducing the amount of data that needs to be transmitted over the network. This feature is particularly useful for utilizing the high sampling rates of PMUs, as it can reduce communication delays in PMU measurement delivery that occur in centralized SE implementations.

GNNs are being applied in a variety of prediction tasks in the field of power systems, including fault location \cite{chen2020FaultLocation}, stability assessment \cite{ZHANG2022Stability}, and load forecasting \cite{ARASTEHFAR2022LoadFOrecasting}. GNNs have also been used for power flow problems, both in a supervised \cite{bolz2019PFapproximator} and an unsupervised \cite{LOPEZ2023GNNpowerFlow} manner. A hybrid nonlinear SE approach \cite{Yang2022RobustPSSEDataDrivenPriors} combines a model and data-based approach using a GNN that outputs voltages which are used a regularization term in the SE loss function.

\textbf{Contributions}:
In our previous work \cite{kundacina2022state}, we proposed a data-driven linear PMU-only state estimator based on GNNs applied over factor graphs. The model demonstrated good approximation capabilities under normal operating conditions and performed well in unobservable and underdetermined scenarios. This work significantly extends our previous work in the following ways:
\begin{itemize}
    \item We conduct an empirical analysis to investigate how the same GNN architecture could be used for power systems of various sizes. We assume that the local properties of the graphs in these systems are similar, leading to local neighbourhoods with similar structures which can be represented using the same embedding space size and the same number of GNN layers.
    \item To evaluate the sample efficiency of the GNN model, we run multiple training experiments on different sizes of training sets. Additionally, we assess the scalability of the model by training it on various power system sizes and evaluating its accuracy, training convergence properties, inference time, and memory requirements.
    \item As a side contribution, the proposed GNN model is tested in scenarios with high measurement variances, using which we simulate phasor misalignments due to communication delays, and the results are compared with linear WLS solutions of SE.
\end{itemize}

\section{Linear State Estimation with PMUs}
\label{sec:lse}
The SE algorithm has a goal of estimating the values of the state variables $\mathbf{x}$, so that they are consistent with measurements, as well as the power system model defined by its topology and parameters. The power system's topology is represented by a graph $\mathcal{G} =(\mathcal{N},\mathcal{E})$, where $\mathcal{N} = {1,\dots,n }$ is the set of buses and $\mathcal{E} \subseteq \mathcal{N} \times \mathcal{N}$ is the set of branches. 

PMUs measure complex bus voltages and complex branch currents, in the form of magnitude and phase angle \cite[Sec.~5.6]{phadke}. PMUs placed at a bus measure the bus voltage phasor and current phasors along all branches incident to the bus \cite{exposito}. The state variables are given as $\mathbf x $ in rectangular coordinates, and therefore consist of real and imaginary components of bus voltages. The PMU measurements are transformed from polar to rectangular coordinate system, since then the SE problem can be formulated using a system of linear equations \cite{kundacina2022state}. The solution to this sparse and noisy system can be found by solving the linear WLS problem: 
\begin{equation}
    \left(\mathbf H^{T} \mathbf \Sigma^{-1} \mathbf H \right) \mathbf x =
		\mathbf H^{T} \mathbf \Sigma^{-1} \mathbf z,    
	\label{SE_system_of_lin_eq}
\end{equation}
where the Jacobian matrix $\mathbf {H} \in \mathbb {R}^{m \times 2n}$ is defined according to the partial first-order derivatives of the measurement functions, and $m$ is the total number of linear equations. The observation error covariance matrix is $\mathbf {\Sigma} \in \mathbb {R}^{m \times m}$, while the vector $\mathbf z \in \mathbb {R}^{m}$ contains measurement values in rectangular coordinate system. The aim of the WLS-based SE is to minimize the sum of residuals between the measurements and the corresponding values that are calculated using the measurement functions \cite{monticelli2000SE}.

This approach has the disadvantage of requiring a transformation of measurement errors (magnitude and angle errors) from polar to rectangular coordinates, making them correlated, resulting in a non-diagonal covariance matrix $\mathbf {\Sigma}$ and increased computational effort. To simplify the calculation, the non-diagonal elements of $\mathbf {\Sigma}$ are often ignored, which can impact the accuracy of the SE \cite{exposito}. We can use the classical theory of propagation of uncertainty to compute variances in rectangular coordinates from variances in polar coordinates \cite{uncertaintyPropagation}. The solution to \eqref{SE_system_of_lin_eq} obtained by ignoring the non-diagonal elements of the covariance matrix $\mathbf {\Sigma}$ to avoid its computationally demanding inversion is referred to as the \textit{approximative WLS SE solution}.

In the rest of the paper, we will explore whether using a GNN model trained with measurement values, variances, and covariances labelled with the exact solutions of \eqref{SE_system_of_lin_eq} leads to greater accuracy compared to the approximative WLS SE, which ignores covariances. The GNN model, once trained, scales linearly with respect to the number of power system buses, allowing for lower computation time compared to both the approximate and exact solvers of \eqref{SE_system_of_lin_eq}.

\section{Methods}

In this section, we introduce spatial GNNs on a high-level and describe how can they be applied to the linear SE problem.

\subsection{Spatial Graph Neural Networks}
\label{sec:gnn}
\begin{figure}[!t]
\centering

\begin{tikzpicture} [scale=0.74, transform shape]

\tikzset{
    box/.style={draw, fill=Goldenrod, minimum width=1.5cm, minimum height=0.8cm}}
    
\begin{scope}[local bounding box=graph]

\node [box]  (message1) at (-0.5, 1.5) {Message};
\node [box]  (message2) at (-0.5, 0) {Message};
\node [below of=message2, font=\Huge, rotate=90] {...};
\node [box]  (messageN) at (-0.5, -2.5) {Message};
\node [box]  (gat) at (3, 0) {$\aggregateFN$};
\node [box]  (update) at (6.3, 0) {$\updateFN$};


\draw[-stealth] (-1.8, 0) -- (message2.west) node[at start,left]{$\mathbf{h_{2}}^{k-1}$};

\draw[-stealth] (-1.8, 1.5) -- (message1.west) node[at start,left]{$\mathbf{h_{1}}^{k-1}$};

\draw[-stealth] (-1.8, -2.5) -- (messageN.west) node[at start,left]{$\mathbf{h_{n_j}}^{k-1}$};

\draw[-stealth] (message1.east) -- (gat.west) node[near start,right]{$\mathbf{m_{1,j}}^{k-1}$};

\draw[-stealth] (message2.east) -- (gat.west) node[midway,below]{$\mathbf{m_{2,j}}^{k-1}$};

\draw[-stealth] (messageN.east) -- (gat.west) node[near start,right]{$\mathbf{m_{{n_j},j}}^{k-1}$};

\draw[-stealth] (gat.east) -- (update.west) node[midway,above]{$\mathbf {m_j}^{k-1}$};

\draw[-stealth] (update.east) -- (7.6, 0) node[at end,right]{$\mathbf{h_{j}}^{k}$};

\draw[-stealth] (6.3, 1.0) -- (update.north) node[at start,above]{$\mathbf {{h_{j}}}^{k-1}$};

\end{scope}
\end{tikzpicture}

\caption{A GNN layer, which represents a single message passing iteration, includes multiple trainable functions, depicted as yellow rectangles. The number of first-order neighbours of the node $j$ is denoted as $n_j$.}
    \label{GNNlayerDetails}
\end{figure}
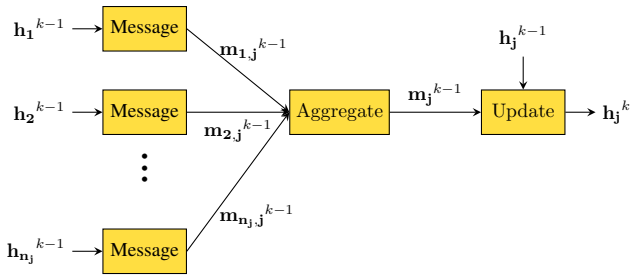

Spatial GNNs are a type of machine learning models that process graph-structured data by iteratively applying message passing to local subsets of the graph. The goal of GNNs is to transform the inputs from each node and its connections into a higher-dimensional space, creating a $s$-dimensional vector $\mathbf h \in \mathbb {R}^{s}$ for each node. GNNs contain $K$ layers, with each layer representing a single iteration $k$ of the message passing process. Each GNN layer includes trainable functions, which are implemented as neural networks, such as a message function, an aggregation function, and an update function, as shown in Fig.~\ref{GNNlayerDetails}. 

The message function calculates the message $\mathbf m_{i,j} \in \mathbb {R}^{u}$ between two node embeddings, the aggregation function combines the incoming messages in a specific way, resulting in an aggregated message $\mathbf {m_j} \in \mathbb {R}^{u}$, and the update function calculates the update to each node's embedding. The message passing process is repeated a fixed number of times, with the final node embeddings passed through additional neural network layers to generate predictions. GNNs are trained by optimizing their parameters using a variant of gradient descent, with the loss function being a measure of the distance between the ground-truth values and the predictions.

\subsection{State Estimation using Graph Neural Networks}
\label{subsec:gnnBasedSe}
The proposed GNN model is designed to be applied over a graph with a SE factor graph topology \cite{cosovic2019bpse}, which consists of factor and variable nodes with edges between them. The variable nodes are used to create a $s$-dimensional embedding for the real and imaginary parts of the bus voltages, which are used to generate state variable predictions. The factor nodes serve as inputs for measurement values, variances, and covariances. Factor nodes do not generate predictions, but they participate in the GNN message passing process to send input data to their neighbouring variable nodes. To improve the model's representation of a node's neighbourhood structure, we use binary index encoding as input features for variable nodes. This encoding allows the GNN to better capture relationships between nodes and reduces the number of input neurons and trainable parameters, as well as training and inference time, compared to the one-hot encoding used in \cite{kundacina2022state}. The GNN model can be applied to various types and quantities of measurements on both power system buses and branches, and the addition or removal of measurements can be simulated by adding or removing factor nodes. In contrast, applying a GNN to the bus-branch power system model would require assigning a single input vector to each bus, which can cause problems such as having to fill elements with zeros when not all measurements are available and making the output sensitive to the order of measurements in the input vector.

Connecting the variable nodes in the $2$-hop neighbourhood of the factor graph topology significantly improves the model's prediction quality in unobservable scenarios \cite{kundacina2022state}. This is because the graph remains connected even when simulating the removal of factor nodes (e.g., measurement loss), which allows messages to be propagated in the entire $K$-hop neighbourhood of the variable node. This allows for the physical connection between power system buses to be preserved when a factor node corresponding to a branch current measurement is removed.

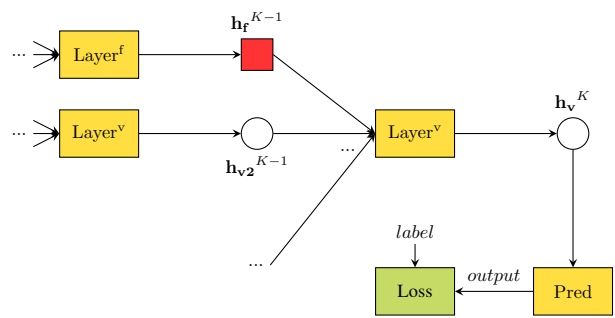
\begin{figure}[!t]
\centering
\begin{tikzpicture} [scale=0.7, transform shape]

\tikzset{
    varNode/.style={circle,minimum size=6mm,fill=white,draw=black},
    factorVoltage/.style={draw=black,fill=red!80, minimum size=6mm},
    factorCurrent/.style={draw=black,fill=orange!80, minimum size=6mm},
    box/.style={draw, fill=Goldenrod, minimum width=1.5cm, minimum height=0.9cm}}

\begin{scope}[local bounding box=graph]

\node [box]  (layerF0) at (-3, 1.5) {$\layerf$};
\node [box]  (layerV0) at (-3, 0) {$\layerv$};

\node[factorVoltage, label=above:$\mathbf{h_{f}}^{K-1}$] (facReV1) at (0, 1.5) {};
\node[varNode, label=below:$\mathbf{h_{v2}}^{K-1}$] (varImV1) at (0, 0) {};
\node[varNode, label=above:$\mathbf{h_{v}}^{K}$] (varReV1) at (6, 0) {};

\node [box]  (layerV1) at (3, 0) {$\layerv$};
\node [box]  (pred) at (6, -3) {$\pred$};
\node [box, fill=SpringGreen]  (loss) at (3, -3) {Loss};

\draw[-stealth] (-4.25, 1.75) -- (layerF0.west);
\draw[-stealth] (-4.25, 1.5) -- (layerF0.west) node[at start,left]{$...$};
\draw[-stealth] (-4.25, 1.25) -- (layerF0.west);

\draw[-stealth] (-4.25, 0.25) -- (layerV0.west);
\draw[-stealth] (-4.25, 0) -- (layerV0.west) node[at start,left]{$...$};
\draw[-stealth] (-4.25, -0.25) -- (layerV0.west);

\draw[-stealth] (layerF0.east) -- (facReV1.west);
	
\draw[-stealth] (layerV0.east) -- (varImV1.west);
	
\draw[-stealth] (varImV1.east) -- (layerV1.west);
	
\draw[-stealth] (0.25, -2.5) -- (layerV1.west)
	node[at start,left]{$...$} node[very near end,left]{$...$};
	
\draw[-stealth] (facReV1.east) -- (layerV1.west);
	
\draw[-stealth] (layerV1.east) -- (varReV1.west);
	
\draw[-stealth] (varReV1.south) -- (pred.north);
	
\draw[-stealth] (pred.west) -- (loss.east)
	node[midway,above]{$output$};
	
\draw[-stealth] (3, -2.1) -- (loss.north)
	node[at start,above]{$label$};

\end{scope}

\end{tikzpicture}


    






	
	
	





\caption{Proposed GNN architecture for heterogeneous augmented factor graphs. Variable nodes are represented by circles and factor nodes are represented by squares. The high-level computational graph begins with the loss function for a variable node, and the layers that aggregate into different types of nodes have distinct trainable parameters.}
    \label{computationalGraph}
\end{figure}

The proposed GNN for a heterogeneous graph has two types of layers: one for factor nodes and one for variable nodes. These layers, denoted as $\layerf$ and $\layerv$, have their own sets of trainable parameters, which allow them to learn their message, aggregation, and update functions separately. Different sets of trainable parameters are used for variable-to-variable and factor-to-variable node messages. Both GNN layers use two-layer feed-forward neural networks as message functions, single layer neural networks as update functions, and the attention mechanism \cite{velickovic2018graph} in the aggregation function. Then, a two-layer neural network $\pred$ is applied to the final node embeddings $\mathbf h^K$ of variable nodes only, to create state variable predictions. The loss function is the mean-squared error (MSE) between the predictions and the ground-truth values, calculated using variable nodes only. All trainable parameters are updated via gradient descent and backpropagation over a mini-batch of graphs. The high-level computational graph of the GNN architecture specialized for heterogeneous augmented factor graphs is depicted in Figure~\ref{computationalGraph}.

The proposed model uses an inference process that requires measurements from the $K$-hop neighbourhood of each node, allowing for computational and geographical distribution. Additionally, since the node degree in the SE factor graph is limited, the computational complexity for the inference process is constant. As a result, the overall GNN-based SE has a linear computational complexity, making it efficient and scalable for large networks.

\section{Numerical Results}
In this section, we conduct numerical experiments to investigate the scalability and sample efficiency of the proposed GNN approach. By varying the power system and training set sizes, we are able to assess the model's memory requirements, prediction speed, and accuracy and compare them to those of traditional SE approaches. 

We use the IEEE 30-bus system, the IEEE 118-bus system, the IEEE 300-bus system, and the ACTIVSg 2000-bus system \cite{ACTIVSg2000}, with measurements placed so that measurement redundancy is maximal. For the purpose of sample efficiency analysis, we create training sets containing 10, 100, 1000, and 10000 samples for each of the mentioned power systems. Furthermore, we use validation and test sets comprising 100 samples. These datasets are generated by solving the power flow problem using randomly generated bus power injections and adding Gaussian noise to obtain the measurement values. All the data samples were labelled using the traditional SE solver. An instance of the GNN model is trained on each of these datasets.


In contrast to our previous work, we use higher variance values of $5 \times 10^{-1}$ to examine the performance of the GNN algorithm under conditions where input measurement phasors are unsynchronized due to communication delays \cite{zhao2019power}. While this is usually simulated by using variance values that increase over time, as an extreme scenario we fix the measurement variances to a high value.

In all the experiments, the node embedding size is set to $64$, and the learning rate is $4\times 10^{-4}$. The minibatch size is $32$, and the number of GNN layers is $4$. We use the ReLU activation function and a gradient clipping value of $5 \times 10^{-1}$. The optimizer is Adam, and we use mean batch normalization.

\subsection{Properties of Power System Augmented Factor Graphs }

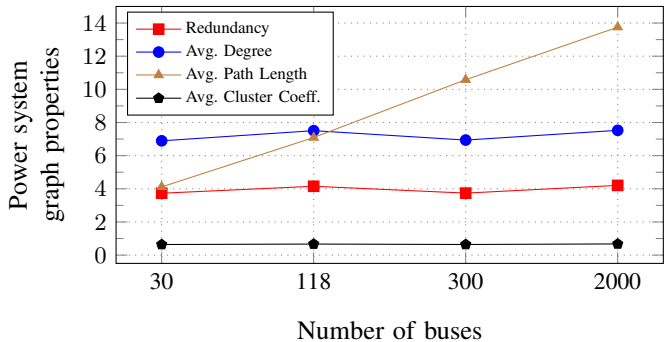
\begin{figure}[!t]
\centering
\begin{tikzpicture}
    \begin{axis} [
        grid=major,
        height=5cm,
        width=\linewidth,
        yscale=1,
        xscale=1,
        xlabel={Number of buses},
        ylabel style={align=center},
        ylabel={Power system\\graph properties},
        xlabel style={
            yshift={-5}, 
        },
        ymin=-0.5, ymax=15,
        ytick={0, 2, 4, 6, 8, 10, 12, 14},
        minor y tick num={1},
        xtick={0, 1, 2, 3},
        xticklabels={30, 118, 300, 2000},
        legend style={
            nodes={
                scale=0.8,
            },
            font=\footnotesize,
            anchor={north west},  
            at={(0.02,0.98)},     
            cells={anchor=west}, 
        },
        legend entries={Redundancy, Avg. Degree, Avg. Path Length, Avg. Cluster Coeff.},
        cycle list={
            {red, mark=square*},
            {blue, mark=*},
            {brown, mark=triangle*},
            {black, mark=pentagon*},
        }
    ]

    \addplot+ coordinates {
        (0, 3.733)
        (1, 4.153)
        (2, 3.740)
        (3, 4.206)
    };
    \addplot+ coordinates {
        (0, 6.894)
        (1, 7.507)
        (2, 6.940)
        (3, 7.527)
    };
    \addplot+ coordinates {
        (0, 4.114)
        (1, 7.093)
        (2, 10.581)
        (3, 13.751)
    };
    \addplot+ coordinates {
        (0, 0.639)
        (1, 0.666)
        (2, 0.641)
        (3, 0.674)
    };
    \end{axis}
\end{tikzpicture}
\caption{Properties of augmented factor graphs along with the system's measurement redundancy for different test power systems, labelled with their corresponding number of buses.}
\label{PowerSystemProperties}
\end{figure}

For all four test power systems, we create augmented factor graphs using the methodology described in Section~\ref{subsec:gnnBasedSe}. Fig.~\ref{PowerSystemProperties} illustrates how the properties of the augmented factor graphs, such as average node degree, average path length, average clustering coefficient, along with the system's maximal measurement redundancy, vary across different test power systems.

The average path length is a property that characterizes the global graph structure, and it tends to increase as the size of the system grows. However, as a design property of high-voltage networks, the other graph properties such as the average node degree, average clustering coefficient, as well as maximal measurement redundancy do not exhibit a clear trend of change with respect to the size of the power system. This suggests that the structures of local, $K$-hop neighbourhoods within the graph are similar across different power systems, and that they contain similar factor-to-variable node ratio. Consequently, it is reasonable to use the same GNN architecture (most importantly, the number of GNN layers and the node embedding size) for all test power systems, regardless of their size. In this way, the proposed model achieves scalability, as it applies the same set of operations to the local, $K$-hop neighbourhoods of augmented factor graphs of varying sizes without having to adapt to each individual case.

\subsection{Training Convergence Analysis}
First, we analyse the training process for the IEEE 30-bus system with four different sizes of the training set. As mentioned in \ref{subsec:gnnBasedSe}, the training loss is a measure of the error between the predictions and the ground-truth values for data samples used in the training process. The validation loss, on the other hand, is a measure of the error between the predictions and the ground-truth values on a separate validation set. In this analysis, we used a validation set of 100 samples.

\begin{figure}[!t]
\centering
\begin{tikzpicture}
    \begin{axis} [
        grid=major,
        yscale=1,
        xscale=1,
        height=5cm,
        width=\linewidth,
        ymin=-0.005, ymax=0.11,
        xlabel={Epoch},
        ylabel={Validation loss},
        minor y tick num={1},
        xtick={1, 30, 60, 90, 120, 150},
        yticklabels={0, 0, 0.05, 0.1},
        xticklabel style={rotate=0},
        yticklabel style={
            /pgf/number format/.cd={
                fixed, 
                fixed zerofill,
                precision=2,
            },
        },
        scaled y ticks=false,
        unbounded coords=jump,
        legend style={
            anchor={north west},
            at={(0.2, 0.98)},
            font=\footnotesize,
            cells={anchor=west},    
            nodes={scale=0.9,},
            legend columns=2,
        },
        cycle list={
            {red, mark=square*},
            {blue, mark=*},
            {brown, mark=triangle*},
            {black, mark=pentagon*},
        },
        legend entries={$n=10$, $n=10^2$, $n=10^3$, $n=10^4$},
    ]

    \addplot+ [
        smooth,
        each nth point={30},
    ] table [
        x=n, y=validation_loss
    ]{figures/fig2/data/150/10_sample_losses.txt};

    \addplot+ [
        smooth,
        each nth point={30},
    ] table [
        x=n, y=validation_loss
    ]{figures/fig2/data/150/100_sample_losses.txt};

    \addplot+ [
        smooth,
        each nth point={30},
    ] table [
        x=n, y=validation_loss
    ]{figures/fig2/data/150/1000_sample_losses.txt};

    \addplot+ [
        smooth,
        each nth point={30},
    ] table [
        x=n, y=validation_loss
    ]{figures/fig2/data/150/10000_sample_losses.txt};
    \end{axis}
\end{tikzpicture}
\caption{Validation losses for trainings on four different training set sizes.}
\label{ieee30Convergence}
\end{figure}
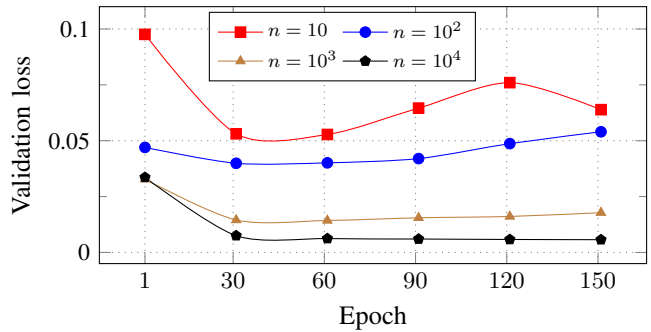

The training losses for all the training processes converged smoothly, so we do not plot them for the sake of clarity. Figure \ref{ieee30Convergence} shows the validation losses for 150 epochs of training on four different training sets. For smaller training sets, the validation loss decreases initially but then begins to increase, which is a sign of overfitting. In these cases, a common practice in machine learning is to select the model with the lowest validation loss value. As it will be shown in \ref{subsec:accuracy}, the separate test set results for models created using small training sets are still satisfactory. As the number of samples in the training set increases, the training process becomes more stable. This is because the model has more data to learn from and is therefore less prone to overfitting.

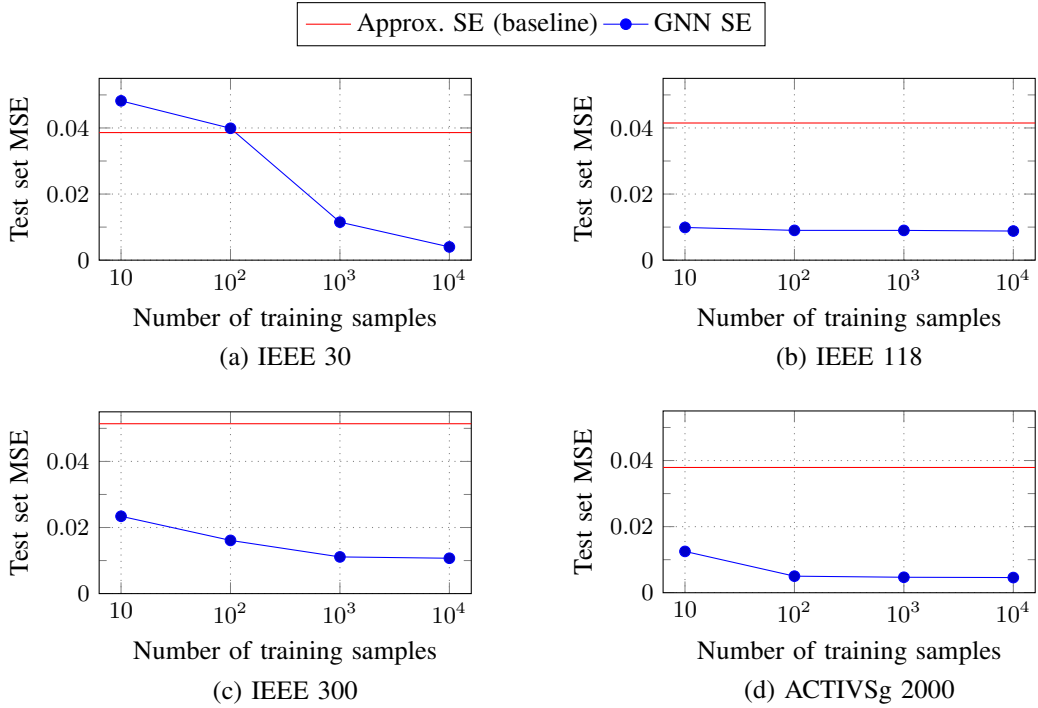
\begin{figure*}[!t]
    \hspace*{\fill}
    \begin{subfigure}[t]{\textwidth}
    \centering
        \begin{tikzpicture}
            \begin{customlegend}[legend entries={Approx. SE (baseline), GNN SE}, legend columns=2,]
                \addlegendimage{draw=red,mark=none,solid,line legend};
                \addlegendimage{blue,mark=*,solid};
            \end{customlegend}
        \end{tikzpicture}
    \end{subfigure}
    \hspace*{\fill}
    \newline
    \hspace*{\fill}
    \begin{subfigure}[t]{0.4\linewidth}
        \centering
        \begin{tikzpicture}
            \begin{axis} [
                title=(a) IEEE 30,
                title style={
                    anchor={north},
                    at={(0.5,-0.5)},
                    yshift=-1,
                },
                grid=major,
                yscale=1,
                xscale=1,
                width=0.9\linewidth,
                height=4cm,
                ymin=0, ymax=0.055,
                xmin=0.8, xmax=4.2,
                xtick=data,
                xlabel={Number of training samples},
                xticklabels={$10$, $10^{2}$, $10^{3}$, $10^{4}$},
                yticklabels={$0$, $0$, $0.02$, $0.04$,},
                ylabel={Test set MSE},
                minor y tick num={1},
                yticklabel style={
                    /pgf/number format/.cd={
                        fixed, 
                        fixed zerofill,
                        precision=2,
                    },
                },
                scaled y ticks=false,
            ]
            \addplot+ [
                color=blue,
            ] table [x=n, y=gnnmsse] {figures/fig4/data/ieee30.txt};
        
            \addplot+ [
                color=red,
                mark=none,
                domain=0.8:4.2,
            ] {0.0386};
            \end{axis}
        \end{tikzpicture}
        \label{subf:ieee30}
    \end{subfigure}
    \hspace{0cm} 
    \begin{subfigure}[t]{0.4\linewidth}
        \centering
        \begin{tikzpicture}
            \begin{axis} [
                title=(b) IEEE 118,
                title style={
                    anchor={north},
                    at={(0.5,-0.5)},
                    yshift=-1,
                },
                grid=major,
                yscale=1,
                xscale=1,
                width=0.9\linewidth,
                height=4cm,
                ymin=0, ymax=0.055,
                xmin=0.8, xmax=4.2,
                ylabel={Test set MSE},
                xlabel={Number of training samples},
                xtick=data,
                xticklabels={$10$, $10^{2}$, $10^{3}$, $10^{4}$},
                yticklabels={$0$, $0$, $0.02$, $0.04$,},
                minor y tick num={1},
                yticklabel style={
                    /pgf/number format/.cd={
                        fixed, 
                        fixed zerofill,
                        precision=2,
                    },
                },
                scaled y ticks=false,
            ]
            \addplot+ [
                color=blue,
            ] table [x=n, y=gnnmsse] {figures/fig4/data/ieee118.txt};
        
            \addplot+ [
                color=red,
                mark=none,
                domain=0.8:4.2,
            ] {0.0415};
            \end{axis}
        \end{tikzpicture}
        \label{subf:ieee118}
    \end{subfigure} 
    \hspace*{\fill}
    \newline\newline
    \hspace*{\fill}
    \begin{subfigure}[t]{0.4\linewidth}
        \centering
        \begin{tikzpicture}
            \begin{axis} [
                title=(c) IEEE 300,
                title style={
                    anchor={north},
                    at={(0.5,-0.5)},
                    yshift=-1,
                },
                grid=major,
                width=0.9\linewidth,
                height=4cm,
                yscale=1,
                xscale=1,
                width=0.9\linewidth,
                height=4cm,
                ymin=0, ymax=0.055,
                xmin=0.8, xmax=4.2,
                xlabel={Number of training samples},
                xtick={1,2,3,4},
                xticklabels={$10$, $10^{2}$, $10^{3}$, $10^{4}$},
                yticklabels={$0$, $0$, $0.02$, $0.04$,},
                ylabel={Test set MSE},
                minor y tick num={1},
                yticklabel style={
                    /pgf/number format/.cd={
                        fixed, 
                        fixed zerofill,
                        precision=2,
                    },
                },
                scaled y ticks=false,
            ]
            \addplot+ [
                color=blue,
            ] table [x=n, y=gnnmsse] {figures/fig4/data/ieee300.txt};
        
            \addplot+ [
                color=red,
                mark=none,
                domain=0.8:4.2,
            ] {0.0514};
            \end{axis}
        \end{tikzpicture}
        \label{subf:ieee300}
    \end{subfigure}
    \hspace{0cm}
    \begin{subfigure}[t]{0.4\linewidth}
        \centering
        \begin{tikzpicture}
            \begin{axis} [
                title=(d) ACTIVSg 2000,
                title style={
                    anchor={north},
                    at={(0.5,-0.5)},
                    yshift=-1,
                },
                grid=major,
                yscale=1,
                xscale=1,
                width=0.9\linewidth,
                height=4cm,
                ymin=0, ymax=0.055,
                xmin=0.8, xmax=4.2,
                ylabel={Test set MSE},
                xlabel={Number of training samples},
                xtick={1, 2, 3, 4}, 
                xticklabels={$10$, $10^{2}$, $10^{3}$, $10^{4}$},
                yticklabels={$0$, $0$, $0.02$, $0.04$,},
                minor y tick num={1},
                yticklabel style={
                    /pgf/number format/.cd={
                        fixed, 
                        fixed zerofill,
                        precision=2,
                    },
                },
                scaled y ticks=false,
            ]
            \addplot+ [
                color=blue,
            ] table [x=n, y=gnnmsse] {figures/fig4/data/activsg2000.txt};
        
            \addplot+ [
                color=red,
                mark=none,
                domain=0.8:4.2,
            ] {0.0379};
            \end{axis}
        \end{tikzpicture}
        \label{subf:activsg2000}
    \end{subfigure}
    \hspace*{\fill}
    \caption{Test set results for various power systems and training set sizes.}
    \label{mse_all_schemes}
\end{figure*}

Next, in Table \ref{tbl_all_schemesEpochs}, we present the training results for the other power systems and training sets of various sizes. The numbers in the table represent the number of epochs after which either the validation loss stopped changing or began to increase. Similarly to the experiments on the IEEE 30-bus system, the trainings on smaller training sets exhibited overfitting, while others converged smoothly. For the former, the number in the table indicates the epoch at which the validation loss reached its minimum and stopped improving. For the latter, the number in the table represents the epoch when there were five consecutive validation loss changes less than $10^{-5}$.

\begin{table}[htbp]
\caption{Epoch until validation loss minimum for various power systems and training set sizes.}
\begin{center}
\resizebox{8.7cm}{!}{
    \begin{tabular}{ | c | c | c | c | }
        \hline
        \textbf{Power system}  & IEEE 118 & IEEE 300 & ACTIVSg 2000                                                       
        \\ \hline
        \textbf{10 samples}    & $61$     & $400$    & $166$                                                              
        \\ \hline
        \textbf{100 samples}   & $38$     & $84$     & $200$                                                              
        \\ \hline
        \textbf{1000 samples}  & $24$     & $82$     & $49$                                                               
        \\ \hline
        \textbf{10000 samples} & $12$     & $30$     & $15$
        \\ \hline

    \end{tabular}
\label{tbl_all_schemesEpochs}
}
\end{center}
\end{table}

Increasing the size of the training set generally results in a lower number of epochs until the validation loss reaches its minimum. However, the epochs until the validation loss reaches its minimum vary significantly between the different power systems. This could be due to differences in the complexity of the systems or the quality of the data used for training.

\subsection{Accuracy Assessment}
\label{subsec:accuracy}

Fig.~\ref{mse_all_schemes} reports the mean squared errors (MSEs) between the predictions and the ground-truth values on 100-sample sized test sets for all trained models and the approximate WLS SE. These results indicate that even the GNN models trained on small datasets outperform the approximate WLS SE, except for the models trained on the IEEE 30-bus system with 10 and 100 samples. These results suggest that the quality of the GNN model's predictions and the generalization capabilities improve as the amount of training data increases, and the models with the best results (highlighted in bold) have significantly smaller MSEs compared to the approximate WLS SE. While we use randomly generated training sets in this analysis, using carefully selected training samples based on historical load consumption data could potentially lead to even better results with small datasets.



\subsection{Inference Time and Memory Requirements}

\begin{figure}[!t]
\centering
\begin{tikzpicture}
    \begin{axis} [
        grid=major,
        ymode=log,
        width=\linewidth,
        height=4.5cm,
        xscale=1,
        yscale=1,
        xlabel={Number of buses},
        ylabel={Inference time ratio},
        xlabel style={
            yshift={-5}, 
        },
        xtick=data,
        xticklabels={30, 118, 300, 2000},
        minor y tick num={1},
        extra y ticks={2, 64},
        extra y tick labels={},
        ytick={1, 2, 10, 64, 100},
        yticklabels={1, 2, 10, 64, },
    ]

    \addplot+ [
        color=blue,
        mark=*,
    ] table [x expr=\thisrowno{0}, y expr=\thisrowno{2}/\thisrowno{3}] {figures/fig3/data.txt};
    \end{axis}
\end{tikzpicture}
\caption{A ratio of the execution times for WLS SE and GNN SE inference on a test set of 100 samples, as a function of the power system size.}
\label{InferTimeRatio}
\end{figure}
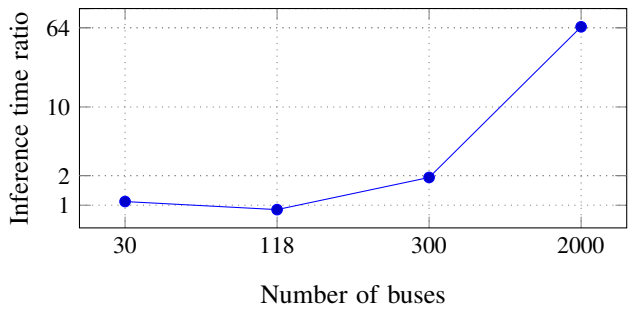

The plot in Fig.~\ref{InferTimeRatio} shows the ratio of execution times between WLS SE and GNN SE inference as a function of the number of buses in the system. These times are measured on a test set of 100 samples. As expected, the difference in computational complexity between GNN, with its linear complexity, and WLS, with more than quadratic complexity, becomes apparent as the number of buses increases. From the results, it can be observed that GNN significantly outperforms WLS in terms of inference time on larger power systems. 






The number of trainable parameters in the GNN model remains relatively constant, as the number of power system buses increases. The number of input neurons for variable node binary index encoding does grow logarithmically with the number of variable nodes. However, this increase is relatively small compared to the total number of GNN parameters\footnote{In fact, all the GNN models we train have the same number of trainable parameters: 49921, which equates to 0.19 MB of memory.}. This indicates that the GNN approach is scalable and efficient, as the model's complexity does not significantly increase with the size of the power system being analysed.

\section{Conclusions}
In this study, we focused on thoroughly testing a GNN-based state estimation algorithm in scenarios with large variances, and examining its scalability and sample efficiency. The results showed that the proposed approach provides good results for large power systems, with lower prediction errors compared to the approximative SE. The GNN model used in this approach is also fast and maintains constant memory usage, regardless of the size of the scheme. Additionally, the GNN was found to be an effective approximation method for WLS SE even with a relatively small number of training samples, particularly for larger power systems, indicating its sample efficiency. Given these characteristics, the approach is worthy of further consideration for real-world applications.

\bibliographystyle{IEEEtran}
\bibliography{cite}

\end{document}